\documentclass[letterpaper, 10 pt, conference]{ieeeconf} 

\IEEEoverridecommandlockouts                             

\usepackage{amsmath,amssymb,amsfonts}
\usepackage{algorithmic}
\usepackage{graphicx}
\usepackage{textcomp}
\usepackage{xcolor}
\usepackage{bm}
\usepackage{lipsum}
\usepackage{booktabs}
\usepackage[style=ieee,dashed=false]{biblatex}

\DeclareSourcemap{
  \maps{
    \map{
      \pertype{article}
      \step[fieldset=language, null]
      \step[fieldset=url, null]
      \step[fieldset=doi, null]
      \step[fieldset=issn, null]
      \step[fieldset=isbn, null]
      \step[fieldset=note, null]
      \step[fieldset=editor, null]
      \step[fieldset=urldate, null]
      \step[fieldset=file, null]
    }
  }
}
\DeclareSourcemap{
  \maps{
    \map{
      \pertype{inproceedings}
      \step[fieldset=language, null]
      \step[fieldset=url, null]
      \step[fieldset=doi, null]
      \step[fieldset=issn, null]
      \step[fieldset=isbn, null]
      \step[fieldset=note, null]
      \step[fieldset=editor, null]
      \step[fieldset=urldate, null]
      \step[fieldset=file, null]
    }
  }
}
\DeclareSourcemap{
  \maps{
    \map{
      \pertype{incollection}
      \step[fieldset=language, null]
      \step[fieldset=url, null]
      \step[fieldset=doi, null]
      \step[fieldset=issn, null]
      \step[fieldset=isbn, null]
      \step[fieldset=note, null]
      \step[fieldset=editor, null]
      \step[fieldset=urldate, null]
      \step[fieldset=file, null]
    }
  }
}

\title{\LARGE \bf
Design and Control of a Multimodal Legged-Aerial Robot Through Separate Control Contraction Metrics
}
\author{
Kaushik Venkatesh Krishnamurthy$^{1}$,
Eric Sihite$^{2}$, Alireza Ramezani$^{1*}$, and Morteza Gharib$^{2}$
\thanks{$^{1}$Authors are with the Silicon Synapse Labs, Department of Electrical and Computer Engineering, Northeastern University, Boston, USA. Emails: 
        {\tt\small 
        venkateshkrishnamu.k, a.ramezani@northeastern.edu}}%
\thanks{$^{2}$Authors are with Center for Autonomus Systems and Technologies, Caltech, Pasadena, CA. Emails: 
        {\tt\small 
        esihite, mgharib@caltech.edu }}%
\thanks{$^{*}$Corresponding author.}%
}

\begin{document}
\maketitle
\begin{abstract}
The integration of dynamic legged locomotion and aerial mobility offers versatility that is deterred by challenges in payload distribution, actuation, and the contrasting requirements of both domains. In this paper, we present the design and operation of a multimodal quadrupedal aerial-legged robot. The presence of legged aerial systems also allows for thruster-assisted legged locomotion, which we make possible with a unified control framework using Control Contraction Metrics (CCM). We search for separate domain-specific contraction metrics for walking and flight independently offline and unify them using an online energy-based Control Lyapunov Function Quadratic Program (CLF QP). This enables optimal, cross-domain control allocation, which we show using simulations. We also show experimental results that validate the mechanical viability of our hardware and the paradigm of thruster-assisted trotting. The stability of said experiments motivates the deployment of the control framework for multimodal operations.
\end{abstract}

\section{Introduction}

Robotic biomimicry of birds' dynamic bipedal and flight locomotion can be a significant ordeal. The prohibitive design restrictions, such as a tight power budget, limited payload, complex multi-modal actuation and perception, an excessive number of active and passive joints involved in each mode, sophisticated control, and environment-specific models, just to name a few, have alienated these concepts. It is worth noting that, as multi-modality has secured birds' (and other species) survival in complex environments, similar manifestations in mobile robots \cite{lock_multi-modal_2013} can be rewarding and important. 

In this work, we will report our legged-aerial quadrupedal robot's first multi-modal operations, including thruster assisted dynamic trotting and hovering. Our long-term objective has remained to integrate dynamic legged locomotion and aerial mobility in a single platform. Our robot design primarily focused on mass allocation to minimize the total cost of transport, which consequently yields smaller actuation bulk in the system, and component selection for the flight unit was systematically made to achieve successful lift-off while not too much negatively affecting dynamic legged mobility.

\begin{figure}
    \centering
    \includegraphics[width=0.8\linewidth]{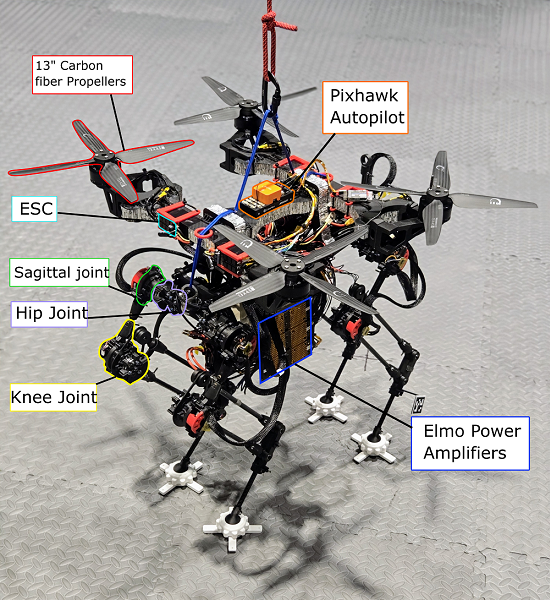}
    \caption{The legged-aerial quadrupedal robot used in this study. The robot has three joint actuator for each leg, and four thrusters for aerial locomotion and thruster-assisted walking.}
    \label{fig:cover}
\end{figure}

\begin{figure}[t]
    \centering \includegraphics[width=\linewidth]{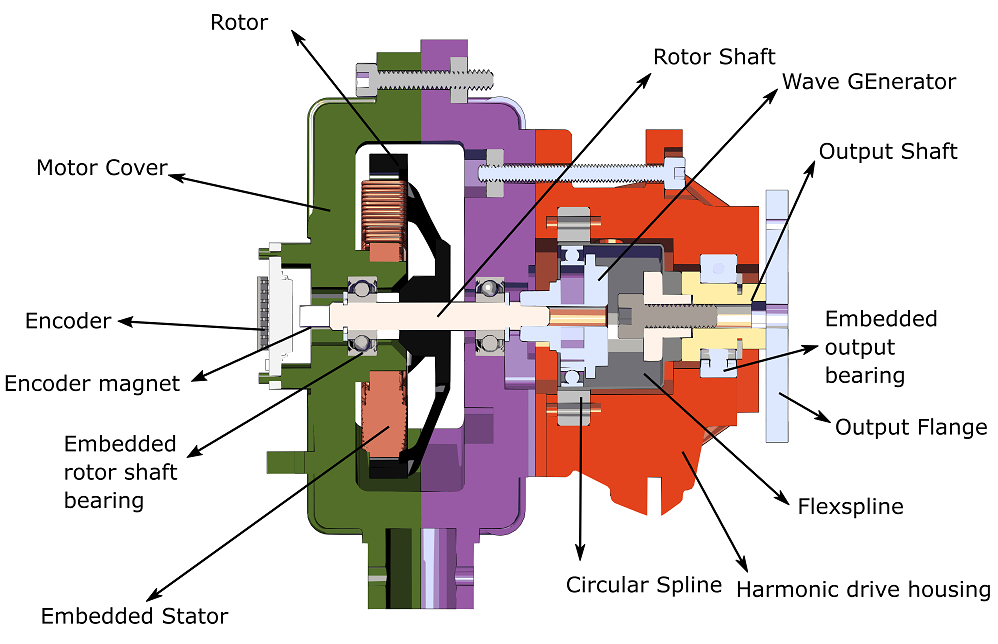}
    \caption{Section view of actuator assembly, showing the harmonic drive, stator, and rotor embedded in the additively manufactured actuator housing.}
    \label{fig:actuator-assembly}
\end{figure}
\begin{table}[t]
    \centering
    \renewcommand{\arraystretch}{1.3} 
    \setlength{\tabcolsep}{10pt} 
    \caption{Robot properties}
    \begin{tabular}{lcc}
        \toprule
        \textbf{Property} & \textbf{Value} & \textbf{Units} \\ 
        \midrule
        Mass & 12 & kg \\
        Height & 0.6 & m \\
        Width & 1 & m \\
        Length & 1.5 & m \\
        Joint Actuator Motor & T-Motor MN4004 Antigravity & - \\
        Motor KV & 270 & kV \\
        Max Motor Torque & 1.8 & Nm \\
        Max Motor Speed & 10000 & rpm \\
        Propeller Motor & MAD motor 5015 & - \\
        Propeller Size & 6.5 & in \\
        Propeller Pitch & 10 & deg \\
        \bottomrule
    \end{tabular}    
    \label{tab:mass-breakdown}
\end{table}

Standard control theory have struggle with dynamic, changing trajectory tracking. A contraction metric is a Riemannian metric, with respect to which the distance between any two solutions of a differential equation is decreasing. Unlike a Lyapunov function, which measure the decreasing generalized distance between a point and an attractor such as an equilibrium, contraction does not require any knowledge about the position of the attractor. Simply, contraction ensures incremental stability between two trajectories, instead of the stability of an equilibrium point. 
Control Contraction Metrics (CCM) ensures that small displacements in directions orthogonal to the span of the control inputs must be contracting. 
The existence of a CCM can be formulated as a convex problem as specified in previous works \cite{manchester_control_2017, singh_robust_2023}. To overcome the computational bottleneck of global stability analysis for this coupled hybrid system, we propose a unified control framework inspired by CCMs. 

\begin{figure*}[t]
    \centering \includegraphics[width = 0.8\linewidth]{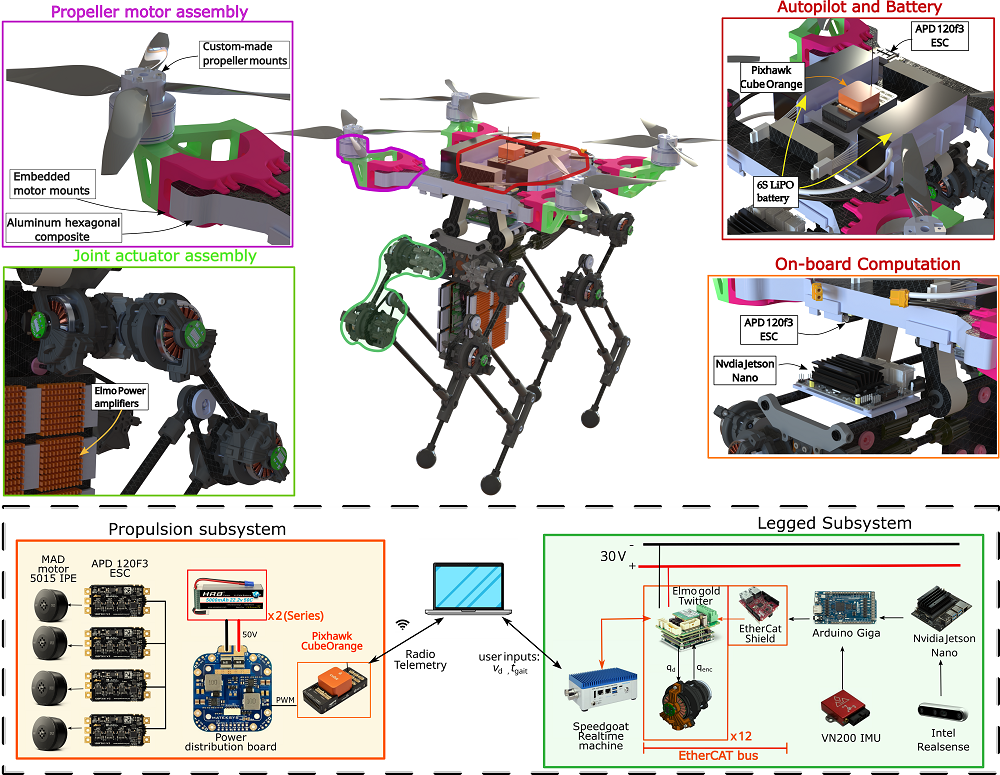}
    \caption{CAD model rendering of the robot, closeup view of all subsystems and components including custom-design actuators, thruster, power electronics, joints, and onboard computers.}
    \label{fig:cad-model}
\end{figure*}

We outline this work's contributions as follows:
\begin{itemize}
    \item The design and operation of a multimodal legged-aerial robot.
    \item A proposed control framework using contraction theory to create separate metrics for walking and flight, which are then unified online using an energy-based control Lyapunov function quadratic programming (CLF-QP).
    \item Experimental results on hardware, with a decoupled heuristic control policy to demonstrate the capabilities of thruster-assisted trotting and flying. These experiments seek to prove that the physical paradigm of combining thrusters with legs is robust.
\end{itemize}
The paper is then structured as follows: we begin with showing the related work with multi-modal robot and CCM, followed by a brief discussion and design choice of the robot's hardware, actuators, and aerial subsystems. Then, we will proceed to the derivations of the CCM synthesis and the online controller, followed by the experimental results and discussions, and ended with the concluding remarks.


\section{Related Work}

{
Multi-modality in a robot expands the robot's capability in performing tasks that a single mode can't do, such as a ground-aerial robot that can use ground locomotion to traverse a relatively flat terrain efficiently and fly over an obstacle when the ground locomotion is unable to drive through it. Therefore, the robot's range and reach can be expanded greatly by utilizing both locomotion modes, and they can augment each other if the design allows it. 

There are various robot designs related to ground-aerial robots, such as, a bipedal robot with quadrotor propulsion attached to its torso \cite{kim_bipedal_2021}, a transforming quadruped that can switch between legged and aerial modes \cite{wang_dynamic_2025}, thruster-assisted hopping robot \cite{haldane_repetitive_2017}, a quadrotor with active wheels underneath \cite{araki2017multi, meiri_flying_2019}, a quadrotor encased in a rolling cage where the thrusters can drive the cage to perform ground locomotion \cite{tagliabue2020shapeshifter}, ornithopters with wheels for ground locomotion \cite{peterson_wing-assisted_2011}, and a transforming robot that can reorient its appendages to transition between a UGV and UAS modes\cite{sihite_multi-modal_2023}. Thrusters can be used to augment ground locomotion capabilities to assist in stabilization by using them to satisfy ground contact constraints \cite{sihite_optimization-free_2021, wang_thruster-enhanced_2025}, can also assist in jumping \cite{sihite_locomotion_2026, krishnamurthy_contact-implicit_2026} and in steep slopes \cite{pitroda_quadratic_2024, krishnamurthy_enabling_2024, krishnamurthy_wair_2026} where is not normally possible using just legged locomotion. We aim to use CCM as a control framework to create metrics for walking and flight to be used in our legged-aerial multi-modal robot.
}

{
There have been many works that use Control Contraction Metrics (CCM) for either Robust or Adaptive Control \cite{manchester_control_2017, manchester_robust_2018, singh_robust_2023, zhao_tube-certified_2022, tsukamoto_learning-based_2021, tsukamoto_neural_2021}. Manchester and Slotine \cite{manchester_control_2017} introduced the idea of Control contraction metrics for a nonlinear system for universal exponential stabilizability. The control contraction metric can then be used for real-time optimization which involves finding a geodesic to the desired state from the current state and finding the control input. This problem is simpler than executing model predictive control. This work then extended to obtain Robust CCMs by using a metric to synthesis a nonlinear state feedback $H\infty$ controller\cite{manchester_robust_2018}. This powerful tool for trajectory, while setting the foundation of our research, assumes exact model knowledge.

To solve control problems at the scale of multicopters and drones, one requires careful structuring of the CCM metric and the polynomial optimization problem to bypass the computational complexity of synthesis. Singh et al. \cite{singh_robust_2023} extended this CCM synthesis with robust feedback control on multicopters and implemented this with experimental results on hardware. This was achieved using exploiting the structure of the problem and Sum of Squares Programming to obtain a metric for a high-dimensional dynamical system. Sun et al. \cite{sun_learning_2021} were able to obtain a tracking controller for different dynamical systems using contraction metrics that were learned using a Neural Network, with controller execution much faster than MPC. Exploring data-driven approaches, Tsukamoto et al. \cite{tsukamoto_neural_2021} introduced Neural Contraction Metrics (NCMs) by training LSTM networks to globally approximate these metrics. The NCM approach derives provably stable, dual estimation and control policies that are strictly optimized for disturbance attenuation. 
Shiromoto et al. \cite{shiromoto_distributed_2019} proposed decentralized control for large networks using contraction theory by forcing a sum-separable global metric, which guarantees stabilization around a target trajectory. Their results demonstrate that utilizing separable structures to break complex global metrics into solvable localized pieces is highly effective. While CCMs have been used to control aerial robots and other simpler systems such as inverted pendulums \cite{manchester_unifying_2018}, we also propose a way to extend this to legged robots. 

\section{Quick Design Overview}

Consumables applied in the metal housing in standard actuators and other supporting components are considered as the bulk of mass in legged robots. Actuators compensate for major external loads, and any design flaw would lead to the concentration of forces internally and consequently yield failures. As a result, the design aims at higher safety factors and often achieve extra hardware security through massive metal housing and support parts. 

To avoid this problem, the fabrication of our robot took place based on embedding electronics and mechanical components precisely at the locations found by previous works. We utilize the MarkForged 3D printer in this section. 
In this process, the circular spline was embedded inside the actuator housings, and other components, including wave generators and flexsplines, were embedded at the subsequent steps. The actuator assembly can be seen in Fig.~\ref{fig:actuator-assembly}.

The benefits of this approach are as follows. First, a considerable payload reduction can be achieved for a feasible and working platform. Second, system components can be precisely located at predetermined locations which are aimed at enhancing efficiency and reducing payload. Last, the model and robot can be matched better, which is important for model-based control design. 
Our legged-aerial quadrupedal robot, shown in Fig.~\ref{fig:cover}, stands at 2.5 ft (0.8 m) tall and is approximately 1 ft (0.3 m) wide. It is fabricated from reinforced thermoplastic materials through additive manufacturing and has a total weight of 12 kg. It hosts on-board power electronics and, currently, is powered using an external power supply. The robot's leg is constructed of four identical legs, where each leg has three active joints (hip frontal, hip sagittal, and knee joints), and the lower leg linkages form a parallelogram mechanism. Therefore, the robot possesses a total of twelve leg actuators.

Each leg actuator is custom-made with a pancake brushless-DC motor winding from T-motor with a KV rating of 400. The motors are actuated by powerful ELMO amplifiers that are located away from the actuator and closer to the body to minimize inertia. These amplifiers are connected on an EtherCAT bus that enables high-speed bidirectional data transfer from a host computer, to command the desired joint position and receive the actual joint position from magnetic joint encoders.

A Harmonic Drive set (flexspline, circular spline, and wave generator) is used for gear reduction with the following gear ratios: 30, 50, and 100 for knee, hip sagittal, and hip frontal, respectively. Some of the power components are embedded into the 3D printed parts, as shown in Fig.~\ref{fig:actuator-assembly}. The current integration has made the robot self-sustained for stable, blind walking and trotting gaits. We note that currently the platform relies on an off-board body orientation sensing system, which soon will be replaced with an on-board inertial measurement unit with negligible weight. 

For flight, the robot is fitted with a frame made of an aluminum hexagonal composite structure sandwiched between carbon fiber plates, which sits on a pair of carbon fiber rails. Additively manufactured motor mounts are embedded in the frame on which the propeller motor is fastened. Custom made hubs house four carbon fiber propellers at a pitch angle of 15 degrees with a diameter of 13 inches. The propeller motors are controlled with APD120F3 ESCs that are powered with two 6S LiPo batteries connected in series for a 12S effective voltage level (approximately 50 volts).

\begin{figure*}[t]
    \centering
    \includegraphics[width=0.75\linewidth]{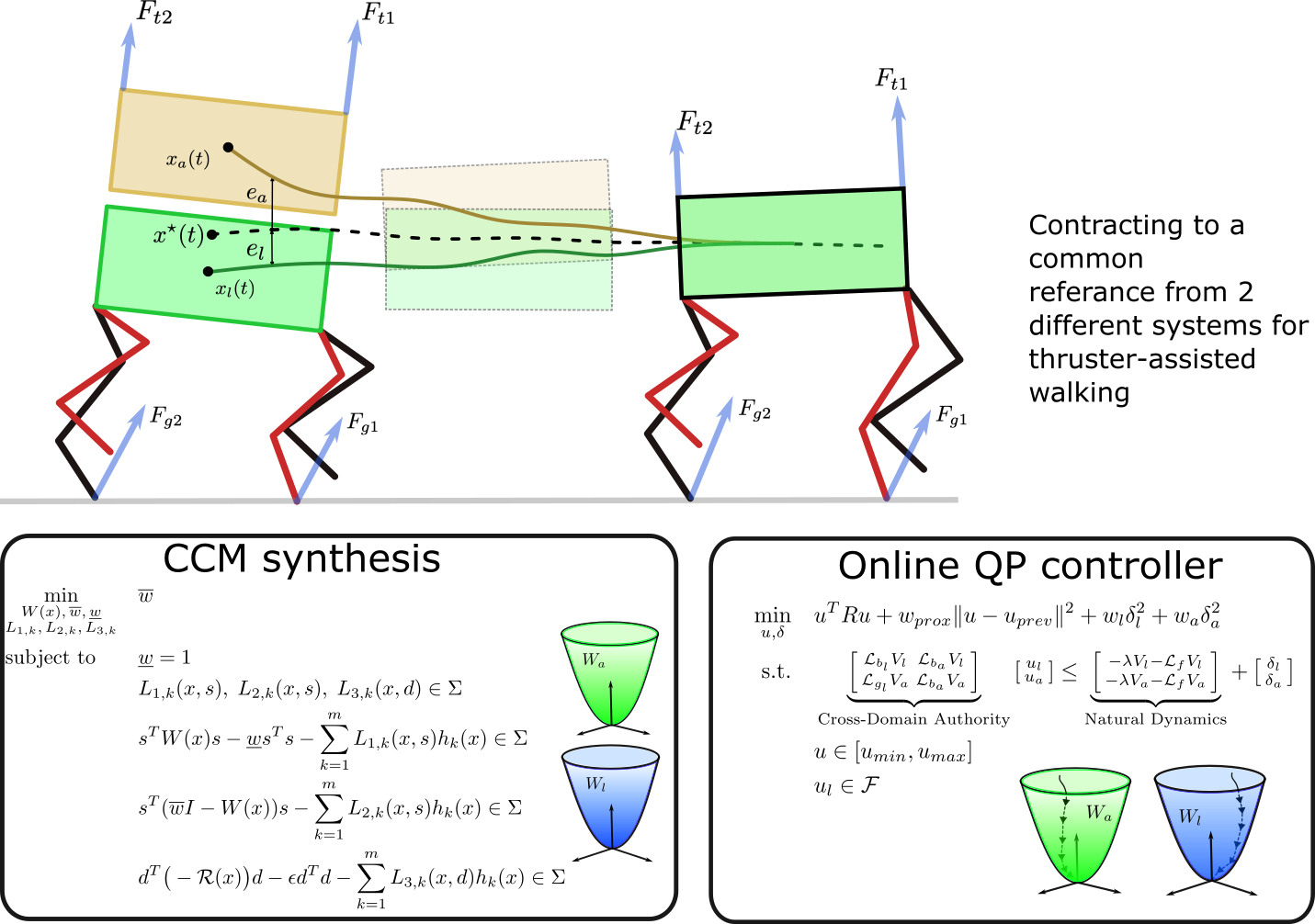}
    \caption{Planar VTOL and legged models used for obtaining contraction metrics. We separate the contraction metrics for the the legged and aerial systems.}
    \label{fig:placeholder}
\end{figure*}

\section{Contraction Metrics}

Incremental stability between 2 trajectories rather than an equilibrium point can be obtained by finding a metric $M$ such that
\begin{equation}
    \dot M  + \frac{\partial f}{\partial x}' M + M\frac{\partial f}{\partial x} \preceq - 2\lambda M,
\end{equation}
then the two trajectories are contracting with rate $\lambda$ for a system with drift dynamics $f$.

Assume a control-affine system, 
\begin{equation}
    \dot x = f(x) + g(x)u,
\end{equation}
It's differential dynamics defined about a point $x$ and $u$, is, 
\begin{equation}
    A = \frac{\partial f}{\partial x} + \frac{\partial g}{\partial x}u.
\end{equation}
The differential dynamics is linear, and we can search for Control Contraction metrics by solving the following
\begin{equation}
    \dot{M} + (A+BK)'M + M(A+BK) \preceq -2\lambda M
\end{equation}
As explained by Singh et al. \cite{singh_robust_2023} and Slotine et al. \cite{manchester_control_2017}, this is a nonlinear inequality and we don't know the metric $M$ and $K$ and becomes hard to solve with convex optimization. We can switch to a dual metric $W = M^{-1}$ and transforming the CCM inequality to, 
\begin{equation}
-\dot{W} + (AW + BKW) + (AW + BKW)' + 2\lambda W \preceq 0
\end{equation}
Taking another variable $Y = KW$, we get, 
\begin{equation}
-\dot{W} + \text{sym}(AW + BY) + 2\lambda W \preceq 0.
\end{equation}
Instead of finding the specific controller $Y$ during the synthesis, we try to determine if a controller exists. We project the dynamics onto the nullspace of the input matrix $B$ and let $B'_{\perp}B=0$. This is important so that the energy in directions that can't be controlled have continuously decreasing energy. 
\begin{align}
    B_\perp&\left( -\dot{W} + AW + WA^T BY + Y^T B^T + 2\lambda W \right) B_\perp \preceq 0 \\
    B_\perp^T & \left( \underbrace{-\dot{W} + AW + WA^T + 2\lambda W }_{\mathcal R} \right) B_\perp \preceq 0.
\end{align}
The time derivative of the metric $W(x)$ can be expanded as 
$$
\dot W = \frac{\partial W}{\partial x}f(x) + \frac{\partial W}{\partial x}Bu.
$$
If the metric depends on the control input $u$ this becomes bilinear $u$ and $W$. To eliminate this, we force the metric to be constant along the control inputs, i.e. 
$$\frac{\partial W}{\partial x}b_j = 0   $$
The conditions that we then solve for are, 
\begin{align}
& C 1: B_{\perp}^T\left(-\dot{W}+A W+W A^T+2 \lambda W\right) B_{\perp} \preceq 0 \\
& C 2: \mathcal{L}_{b_j} W=0,
\label{eq: cond}
\end{align}
We can define the model as follows  
\begin{equation}
    x = \begin{bmatrix}
        p_x \\ p_y \\ \phi \\ v_x \\ v_y \\ \dot{\phi}
    \end{bmatrix}, 
    f(x) = \begin{bmatrix}
        v_x \cos\phi - v_y \sin\phi \\
        v_x \sin\phi + v_y \cos\phi \\
        \dot \phi \\
        v_y \dot \phi - g \sin\phi \\
        -v_x \dot \phi - g \cos\phi \\
        0
    \end{bmatrix}, 
\end{equation}
where $v_x$ and $v_z$ are velocities in the body frame. Assuming the planar VTOL is actuated using 2 body frame fixed forces $F_1$ and $F_2$ which form $u_a$. The legged system is actuated by the effective wrench of two 2D contact forces $F_{g1}$ and $F_{g2}$ which forms $u_l$ . The unactuated dimensions for a legged robot are then $x_{\perp} = [p_x, p_y, \phi]$ and for the planar VTOL, $x_{\perp} = [p_x, p_y, \phi, v_x]$ and the control mapping matrices,
$$
B_{a} = \begin{bmatrix} 0 & 0 \\ 
0 & 0 \\
0 & 0 \\
1/m  &1/m  \\
-l/J & l/J\end{bmatrix}_{6\times 2}  \, , B_{l} = \begin{bmatrix} 0 & 0 & 0\\ 
0 & 0 & 0 \\
0 & 0 & 0 \\
1/m &0 & 0  \\
0   & 1/m & 0 \\
0  &   0  & 1/J 
\end{bmatrix}_{6\times 3} 
$$
where $m$ and $J$ are the mass and the rotational inertia of the body, respectively. The distance between the two thruster positions on the body frame is denoted by $2l$. The perpendicular matrices are then, 
$$
B_{\perp,a} = \begin{bmatrix}
I_{4} \\
0_{4\times 2}
\end{bmatrix} , \quad B_{\perp,l} = \begin{bmatrix}
I_{3} \\
0_{3\times3}
\end{bmatrix}.
$$
Because of the selection of the state dimensions in the body frame velocities, the $B$ and $B_{\perp}$ are constant matrices. This significantly simplifies the synthesis of a contraction metric. The search for a contraction metric then boils down to an optimization problem. The objective is to find a metric that is positive definite, bounded, and that satisfies the contraction conditions C1 and C2~\eqref{eq: cond}.
\begin{equation}
\begin{aligned}
& \underset{W(x), \, \overline{w}, \, \underline{w}}{\text{min}}
& & \overline{w} \\
& \text{s.t.} & & \underline{w} = 1 \\
&&& W(x) - \underline{w} I \succeq 0, \quad \forall x \in \mathcal{X} \\
&&& \overline{w} I - W(x) \succeq 0, \quad \forall x \in \mathcal{X} \\
&&& B_{\perp}'(\mathcal R(x))B_{\perp} \preceq 0, \quad \forall x \in \mathcal{X} \\
&&& \mathcal L_{b_j}W(x) = 0, \quad \forall x, \in \mathcal X,
\end{aligned}
\end{equation}
where $\overline{w}$ and $\underline{w}$ are the upper and lower bounds largest and smaller eigenvalues, respectively and $\mathcal X$ is the set of allowable states. Instead of searching for an infinite number of possible PSD matrices, we assume $W(x)$ as a matrix of polynomials and we enforce that it is positive definite using polynomial optimization. Let us assume $W(x)$ is a weighted sum of a basis of monomials $m_k$
\begin{equation}
    W(x) = \sum_k C_km_k(x)
\end{equation}
We solve for $C_k$ and the monomials are defined from the state variables. The parameterization of the dual metric $W(x)$ is inspired by \cite{singh_robust_2023}, where the trigonometric terms are parametrized using the Chebyshev polynomials defined in the operating regime. We exploit translational invariance by ignoring positions, and to satisfy Condition C2, we force $W(x)$ to be independent of the directly actuated states ($v_z$ and $\dot{\phi}$). To avoid the conservatism of a global metric, we constrain the synthesis to a specific flight envelope with box constraints defined by $h_k(x) \ge 0 , \quad \forall k=1,\dots,m$. Using polynomial multipliers $L_{i,k}$ as an S-procedure relaxation, we enforce the stability criteria strictly within box constraints. The polynomial optimization is formulated as,
%
%
\begin{gather}
\underset{\substack{W(x), \, \overline{w}, \, \underline{w}, \, L_{1,k}, \, L_{2,k}, \, L_{3,k}}}{\text{min}} \overline{w} \hspace{2in} \nonumber \\
\begin{split}
\text{s.t.} \quad & \underline{w} = 1 \\
& L_{1,k}(x, s), \; L_{2,k}(x, s), \; L_{3,k}(x, d) \in \Sigma \\
& s^T( W(x) - \underline{w}I ) s - \sum_{k=1}^m L_{1,k}(x, s) h_k(x) \in \Sigma \\
& s^T (\overline{w} I - W(x)) s - \sum_{k=1}^m L_{2,k}(x, s) h_k(x) \in \Sigma \\
& d^T \big(-\mathcal R(x)\big) d - \sum_{k=1}^m L_{3,k}(x, d) h_k(x) \in \Sigma        
\end{split}
\end{gather}
where, $\Sigma$ is the cone of SOS polynomials. The metric $M$ can then be obtained by inverting $W$. 

\subsection{Domain specific contraction metrics}

Contractions are synthesized by projecting the system dynamics into the unactuated subspace, and these are different for the PVTOL and the legged systems. The legged system with two contact points has one more dimension of actuation, i.e velocity in the fixed $x$-direction. Furthermore, to capture both physics perfectly, a global metric would need to depend on the variables for both systems. The legged metric needs relative height ($y$) to calculate lever arms. The aerial metric needs body velocities ($v_x, v_y$) to handle aerodynamic coupling and rotating thrust vectors. This becomes significantly harder to solve using SOS programming because of the computational complexity. 
Separating the metrics significantly simplifies the synthesis of the contraction metric while still being able to use them with an energy based online controller that evaluates how both sets of actuators impact both metrics. With cross-domain authority explained in the following section, we can see how the QP can find control inputs that help reduce energy in both funnels. Therefore, we make use of 2 different metrics $M_l(x)$ and $M_a(x)$ corresponding to mapping functions $B_l$ and $B_a$ i.e the aerial and legged subsystems.

\begin{figure*}[t]
    \centering
    \includegraphics[width=1\linewidth]{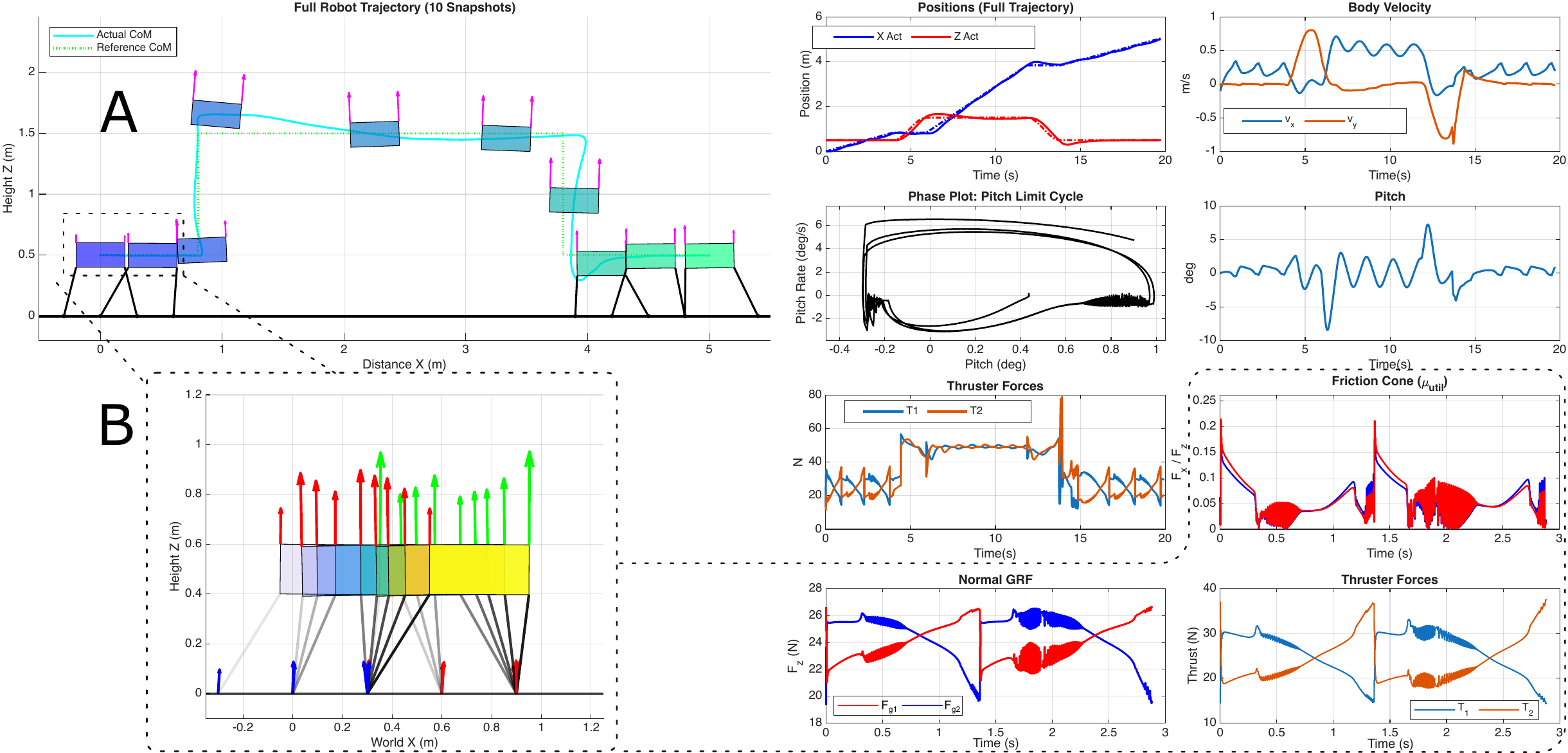}
    \caption{A. Snapshots of Matlab simulation showing thruster-assisted trotting using the min-norm controller with the unified contraction constraints. B. Zoomed in view of a couple of steps, showing the relative direction and magnitude of front and rear thruster forces and contact forces. Along with trajectories of positions, velocity, pitch, phase plots, contact forces, and constraint satisfaction.}
    \label{fig:sim_results}
\end{figure*}

\begin{figure*}[t]
    \centering
    \includegraphics[width=0.88\linewidth]{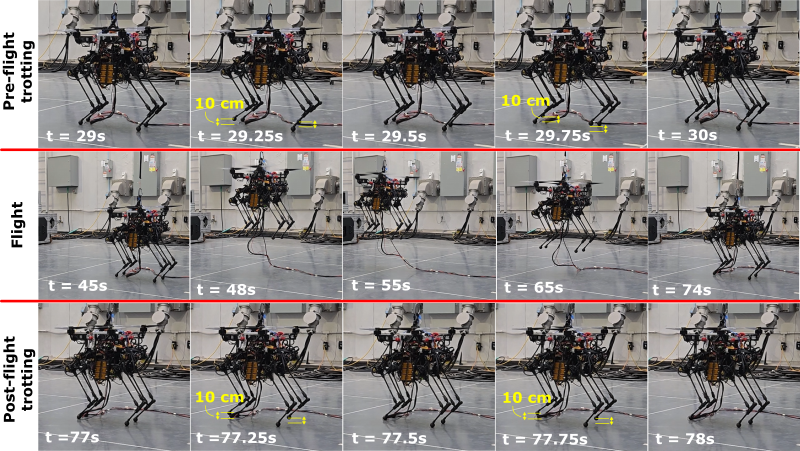}
    \caption{Snapshots from experimental results. The legged phase involves 2-point contact trotting while a human operator provides thruster inputs to. During the flight phase, the flight operator provides thruster inputs through the onboard CubeOrange flight controller. The robot lands again and continues trotting stably. }
    \label{fig: plots}
\end{figure*}

\subsection{Reimannian Energy as a CLF}

Let us define an energy function, 
\begin{equation}
V = e'M(x)e,    
\end{equation}
where $e$ is the Euclidean error to the common reference to track for both systems. A reducing energy then ensures tracking of the reference. This can be ensured by ensuring reducing energy, by $\dot V < - \lambda_u V$, with convergence rate $\lambda_u$.  This can be further expanded as, 
\begin{equation}
\begin{aligned}
\dot V &= \frac{\partial V}{\partial x}(f(x)+g(x)u) \\
&= \mathcal L_fV + \mathcal L_gVu < -\lambda_u V.
\end{aligned}
\end{equation}
Since we are solving for the control inputs, we rearrange the equations so that control input terms are on the left, 
\begin{equation}
    \mathcal{L}_g V u \le -\lambda_u V - \mathcal{L}_f V
\end{equation}
The term $\mathcal{L}_g V u$ can be expanded as, 
\begin{equation}
    \mathcal{L}_g Vu = \begin{bmatrix}
        \mathcal{L}_{g_l}V & \mathcal{L}_{g_a}V
    \end{bmatrix}\begin{bmatrix}
        u_l \\
        u_a
    \end{bmatrix}.
\end{equation}

Since we synthesized two CCMs, we can use them together to ensure that we have reducing energy on both manifolds to track the reference trajectory using control inputs from both systems. By passing both metric derivatives into the QP simultaneously as inequality constraints, we force the solver to find a single, unified control vector $u = [u_{l}, u_{a}]^T$ that lies in the intersection of both stability half-spaces. The reducing energy constraints with the separate metrics can then be formed as, 
$$
\begin{aligned}
\underbrace{\left[\begin{array}{ll}
\mathcal L_{g_{l}}  V_l & \mathcal L_{b_{a}} V_l \\
\mathcal L_{g_{l}}  V_a & \mathcal L_{b_{a}} V_a\end{array}\right]}_{\text {Cross-Domain Authority }}
\left[\begin{array}{l}
u_{l } \\
u_{a }
\end{array}\right] &\leq \underbrace{\left[\begin{array}{l}
-\lambda_u V_l-\mathcal L_f V_l \\
-\lambda_u V_a-\mathcal L_f V_a
\end{array}\right]}_{\text {Natural Dynamics }} \\
Pu &\leq q.
\end{aligned}
$$
This way we can ensure that both sets of control inputs enforce reducing energy on both manifolds. To find the control effort online, we use a min-norm control effort QP, where the constraints enforce a consistent reduction in this energy. This QP is executed in the simulation at every instant,  
\begin{equation}
\begin{aligned}
\min _{u, \delta} \quad  & u^T R u+w_{p r o x}\left\|u-u_{p r e v}\right\|^2+w_L \delta_L^2+w_A \delta_A^2 \\
\text{s.t} \quad & Pu<Q+\left[\begin{array}{l}
\delta_l \\
\delta_a
\end{array}\right] \\
&u \in [u_{min}, u_{max}]\\
&u_{l} \in \mathcal{F} \quad (\text{Friction Cone})
\end{aligned}
\end{equation}

Here, $R \in \mathbb R^{6 \times 6}$ is a weighting matrix that penalizes the individual control inputs, the $w_{prox}$ is a proximal cost to avoid sudden changes between consequent in control input $w_l$. The loss functions also have slack variables $\delta_l$ and $\delta_a$ with weights $w_l$ and $w_a$, respectively. For pure aerial flight or walking, the constraints just reduce to reducing energy of the corresponding system. 
\begin{figure}
    \centering
    \includegraphics[width=\linewidth]{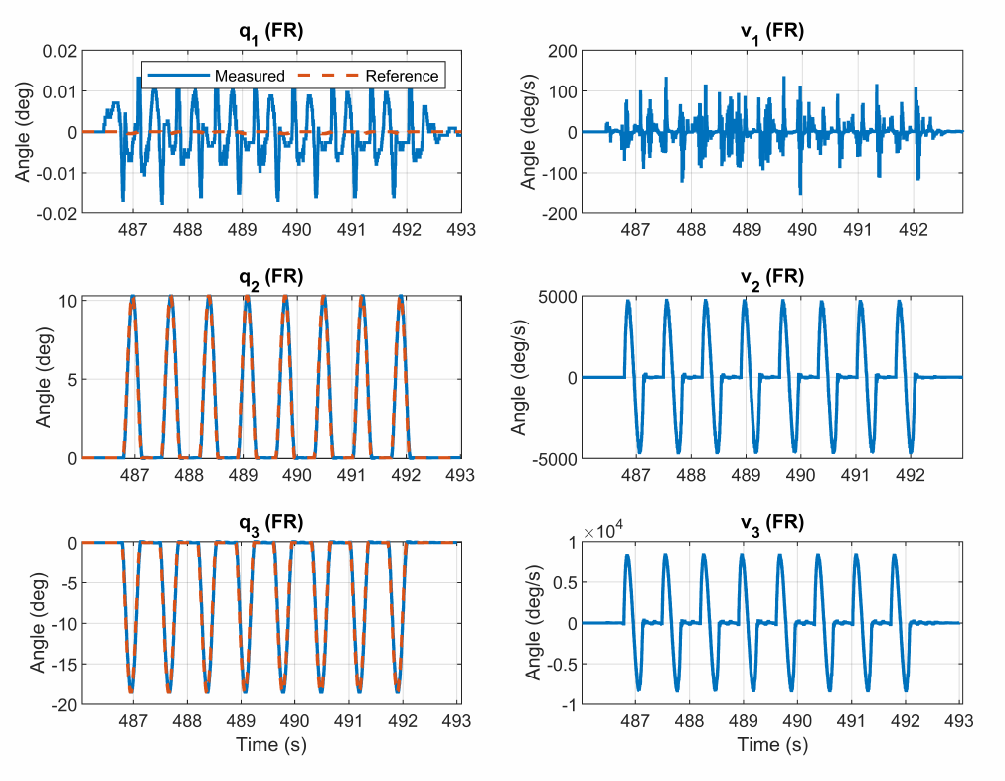}
    \caption{Joint positions and velocities reference and tracking during thruster-assisted trotting experiments}
    \label{fig: plots}
\end{figure}
\section{Results}

\begin{figure}
    \centering
    \includegraphics[width=0.8\linewidth]{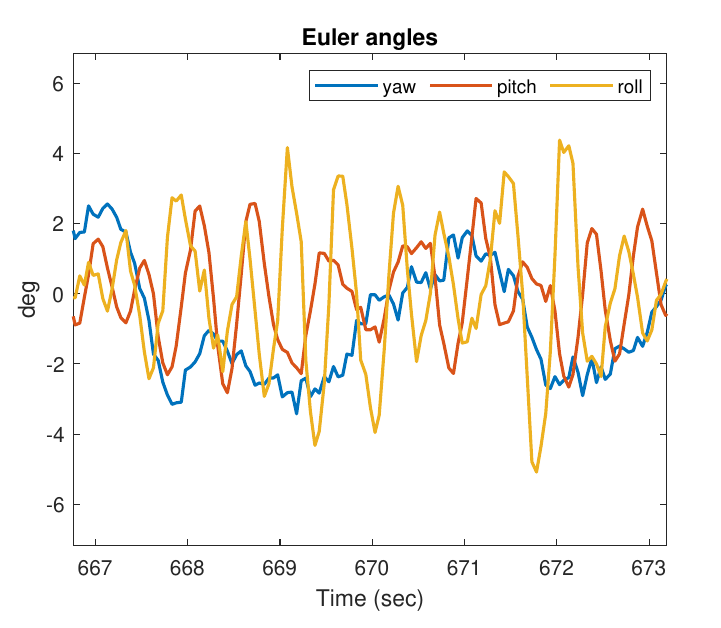}
    \caption{Euler angles in the thruster assisted trotting.}
    \label{fig:enter-label}
\end{figure}

\begin{figure}[t]
    \centering \includegraphics[width=0.95\linewidth]{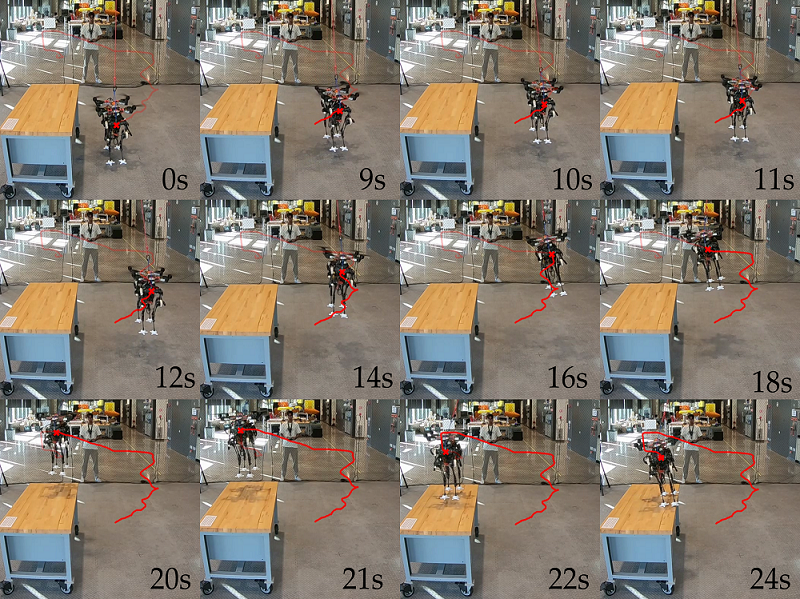}
    \caption{Snapshots of the robot flying and landing on an obstacle.}
    \label{fig: exploding_view}
\end{figure}


\subsection{Simulation }

The CCM synthesis and the robot's simulations were done on Matlab. By assuming significant control authority in the roll directions, we simulated forward walking with fixed contact locations for a two-point contact gait, and performed a reduced-order simulation of the robot to illustrate the effectiveness of this control methodology. The physics simulation runs at 1 kHz, and the control is applied at 250 Hz. 
A forward velocity reference of 0.3 m/s was applied at a height of 0.5 meters from the ground. Foot locations were updated using a Raibert heuristic to track a forward velocity. With these parameters, stable thruster assisted walking was possible. Figure~\ref{fig:sim_results} shows these simulation results with snapshots, over the entire trajectory, along with zoomed-in snapshots to show individual thruster forces, contact forces, and constraint satisfaction. 
We see that the thruster forces and contact forces work collaboratively to move the robot while also respecting contact constraint satisfaction. The simulation shows that we can see a stable limit cycle on the pitch angle during walking. From the simulation, the thruster forces account for about half of the robot's weight. During flight, the constraints for the online QP only ensure a reduction in energy with respect to the aerial metric to track an upward velocity to a desired height, fly forward, and then land again upon which the unified contraction metrics are resumed.

\subsection{Experimental Results}

We tested the ability of the thrusters to stabilize the body dynamics during trotting by comparing thruster-assisted vs thruster unassisted trotting. The propulsion system was controlled using the Pixhawk CubeOrange and controlled by a human operator. The trotting experiment was done with a gait time of 0.5 secs. The legged system is actuated by commanding the motors to track a predefined gait with  From the experiments, the thrusters provide a force that lower the apparent gravity that the robot feels. This allows us to slow down the gait time and gives the legs a much more time to swing into place. Simplistically, the thrusters act as a highly damped spring that suppresses the effects of foot impacts during dynamic walking. This effect can even be seen with a naive PD controller on the aerial subsystem. While these heuristic baseline experiments validate the mechanical viability and inherent robustness of thruster-assisted locomotion, they rely on rigid, domain-isolated control allocation. The naive policy cannot formally guarantee stability during aggressive transient maneuvers, nor can it optimally distribute the control effort when friction limits are approached. This further motivates the need for a unified, optimization-based framework that actively exploits the coupling between the aerial and legged domains.

\section{Concluding Remarks}

In conclusion, here we present the hardware design, system integration, and fabrication for a novel multimodal-legged aerial quadruped robot. A control framework with separate CCMs is proposed along with the synthesis process. We then propose a unified QP that can optimally allocate control effort across both domains. The experimental results on hardware validate the capabilities of the multimodal platform and demonstrate that thruster assistance mitigates and dampens the effects of impacts from dynamic walking. Future work will focus on using the proposed unified CCM-QP on hardware, as well as integrating onboard state estimation to achieve fully autonomous, multi modal maneuvers.

\section{Acknowledgments}
This work was supported by the National Science Foundation CAREER Award (Award number 2340278)

\printbibliography
\end{document}